\pdfoutput=1

\documentclass[11pt]{article}

\usepackage{acl}

\usepackage{times}
\usepackage{csquotes}
\usepackage{latexsym}
\usepackage{multirow}   
\usepackage{graphicx}   
\usepackage{enumitem}
\usepackage{amsmath}    
\usepackage{amssymb}    
\usepackage{latexsym}
\usepackage[inkscapelatex=false]{svg}
\usepackage{url}    
\usepackage{colortbl, booktabs} 
\usepackage{makecell} 
\usepackage{color}  
\usepackage{array}
\usepackage{hyperref}   
\usepackage[all]{nowidow}   
\usepackage[subtle]{savetrees} 

\usepackage{algorithm}
\usepackage{algorithmic}

\usepackage{microtype}
\usepackage{multicol}
\usepackage{tipa}

\usepackage{inconsolata}

\usepackage[T1]{fontenc}

\usepackage[utf8]{inputenc}

\title{\textit{Who's~Laughing~Now?}\ An Overview of Computational Humour Generation and Explanation}

\author{Tyler Loakman\textsuperscript{1},
William Thorne\textsuperscript{1},
Chenghua Lin\textsuperscript{2}\\
  \textsuperscript{1}Department of Computer Science, The University of Sheffield, UK \\
  \textsuperscript{2}Department of Computer Science, The University of Manchester, UK\\
  \texttt{tcloakman1@sheffield.ac.uk} \\
  \texttt{wthorne1@sheffield.ac.uk} \\
  \texttt{chenghua.lin@manchester.ac.uk}}

\begin{document}

\maketitle

\begin{abstract}
The creation and perception of humour is a fundamental human trait, positioning its computational understanding as one of the most challenging tasks in natural language processing (NLP). As an abstract, creative, and frequently context-dependent construct, humour requires extensive reasoning to understand and create, making it a pertinent task for assessing the common-sense knowledge and reasoning abilities of modern large language models (LLMs).
In this work, we survey the landscape of computational humour as it pertains to the generative tasks of creation and explanation. We observe that, despite the task of understanding humour bearing all the hallmarks of a foundational NLP task, work on generating and explaining humour beyond puns remains sparse, while state-of-the-art models continue to fall short of human capabilities. We bookend our literature survey by motivating the importance of computational humour processing as a subdiscipline of NLP and presenting an extensive discussion of future directions for research in the area that takes into account the subjective and ethically ambiguous nature of humour.

\end{abstract}

\section{Introduction}
Humour serves as a foundational element of human communication, acting as a way through which to express emotion, build interpersonal relationships, and experience levity and entertainment \cite{ritschel-andre-2018-shaping}. However, humour may arise from myriad sources \cite{dynel-2009}, including simple, innocuous wordplay such as puns, all the way to deeply contextualised topical references that require layered reasoning to both create and interpret \cite{highfield_2015, Laineste_2002}. The position of humour as a distinctly human experience, in addition to the challenges its processing presents, even to humans \cite{BellAttardo+2010+423+447,MAK_CARPENTER_2007,Wierzbicki01071978}, makes generative computational humour tasks such as joke creation and explanation a formidable domain for analysing the common sense reasoning capabilities of modern large language models (LLMs).

\subsection{Existing Surveys} 
Several surveys have examined different aspects of computational humour. Early foundational work by \citet{ritchie_current_2001} provides a comprehensive overview of the emerging field of computational humour. More recent reviews have focused on specific subdomains of computational humour. \citet{Ramakristanaiah_2021_survey} and \citet{Kalloniatis2024} survey humour recognition and detection, while \citet{kenneth2024systematicliteraturereviewcomputational} review humour style classification. \citet{hindi_humour_survey_2024} explore sarcasm and humour detection in code-mixed Hindi, and \citet{Nijholt2017HumorIH} cover the unique domain of humour in human-computer interaction. The most comparable works to ours are \citet{amin-burghardt-2020-survey}, who present a survey of text-based computational humour generation, and \citet{nguyen-ng-2024-computational}, who survey works on meme understanding. Our survey provides an up-to-date account of the field in the age of LLMs, an additional focus on humour explanation, and suggestions for future work based on real-world ethical considerations, rather than solely technical innovations.

\subsection{Survey Outline}
\label{sec:outline}
This survey is structured around the two primary generative tasks in computational humour. In \S \ref{sec:generation} we explore humour generation, broken down by humour type, whilst in \S \ref{sec:explanation} we explore humour explanation, broken down into explanation through classification (see \S \ref{sec:classification}) and natural language explanation (see \S \ref{sec:natural_language_explanations}). Finally, in \S \ref{sec:discussion} we provide future directions for research in generative computational humour, taking into account the complex ethics of a potentially offensive language form, and the nature of generating creative text.

\section{The Role of Humour Research in NLP}

\begin{figure*}
    \centering
    \includegraphics[width=0.9\linewidth]{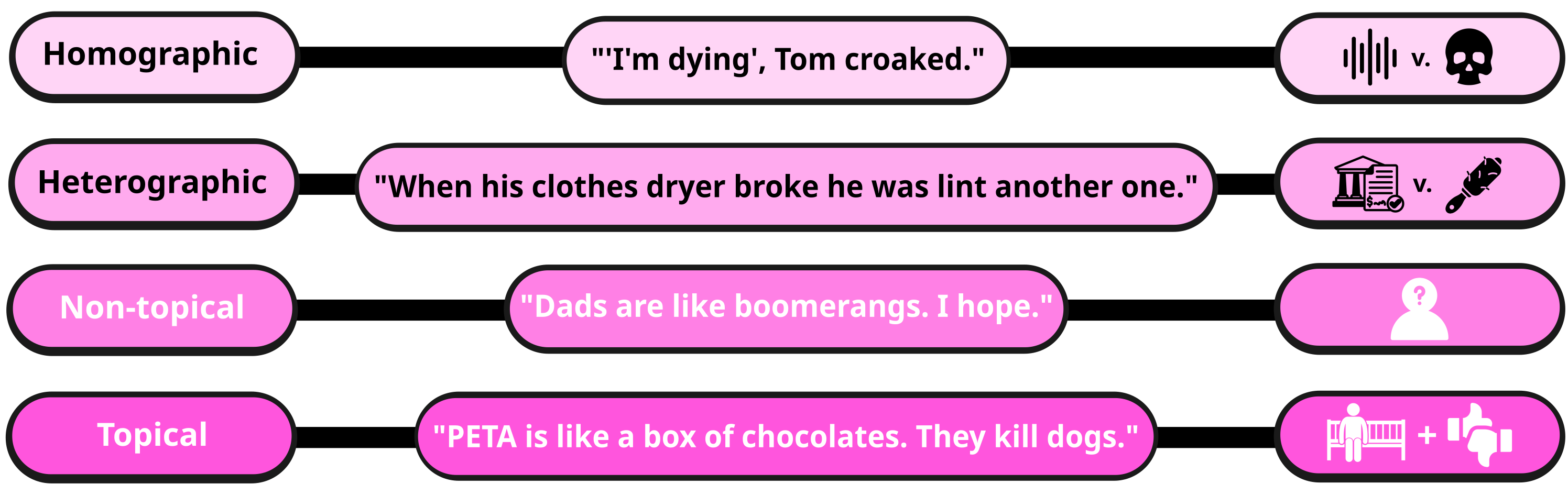}
    \caption{Examples of 4 broad textual joke types from \citet{loakman2025comparingapplesorangesdataset}. Homographic and Heterographic refer to types of puns, whilst Topical and Non-Topical relate to incongruity-based humour, themed around common sense and contemporary news, respectively. The \textit{homographic} pun exploits the dual meaning of "croaked" as both a style of speech and a euphemism for dying; the \textit{heterographic} pun relies on the phonetic similarity between "lint" and "leant"; the \textit{non-topical} joke plays on the trope of an absentee father (not) returning like a boomerang; and the \textit{topical} joke refers to the animal welfare organisation PETA's high euthanasia rates and a reference to the movie Forrest Gump. Each joke type plays on polysemy, phonetics, social constructs, and esoteric knowledge, respectively. }
    \label{fig:text_examples}
\end{figure*}

Significant effort and resources are currently being invested into enhancing the reasoning capabilities of language models \cite{wu-etal-2024-reasoning, yin-etal-2024-reasoning, servantez-etal-2024-chain}. Reasoning or thinking models embed established prompting techniques such as \textit{chain-of-thought} \cite{weiChainofThoughtPromptingElicits2023} or \textit{tree of thought} \cite{yao2023treethoughtsdeliberateproblem} into the generation process, typically producing a verbose stream of consciousness in order to elicit the necessary steps to solve complex tasks. Scaling test-time compute in this way has seen a significant improvement on many benchmarks \cite{geipingScalingTestTimeCompute2025, snellScalingLLMTestTime2024}; however, the current status quo relies heavily on more technical and formalised reasoning tasks such as code generation and arithmetic. Whilst results in such domains can be deterministically validated \cite{wang-etal-2024-math, wang-etal-2024-multi-step}, and large quantities of samples can be generated \cite{xu2025kodcodediversechallengingverifiable}, verbal and common-sense reasoning remains a fundamental task with wider applicability to the end-users of such technologies \cite{trichelair-etal-2019-reasonable}. We argue, therefore, that computational humour processing is an essential task for addressing the limitations in how models process natural language pragmatics and cultural knowledge.

To emphasise the existing focus on mathematical reasoning, of the benchmarks suites used in recent releases (i.e., Kimi K2\footnote{\url{https://moonshotai.github.io/Kimi-K2/}} and Grok4\footnote{\url{https://x.ai/news/grok-4}}), nine focus on STEM problem solving: GPQA \cite{rein2024gpqa}, USAMO \cite{petrov2025proofbluffevaluatingllms}, HMMT,\footnote{\url{https://www.hmmt.org/}} AIME-25,~\footnote{\url{https://artofproblemsolving.com}} LiveCodeBench \cite{jain2024livecodebenchholisticcontaminationfree}, SWE \cite{jimenez2024swebenchlanguagemodelsresolve}, OJBench \cite{wangOJBenchCompetitionLevel2025}; or structured reasoning: ARC-AGI-2 \cite{chollet2025arcagi2newchallengefrontier}, and ACEBench \cite{chen2025acebenchwinsmatchpoint}. On the other hand, only Tau2 \cite{barres2025tau2benchevaluatingconversationalagents} touched upon communicative competence, albeit through customer service scenarios. 

Computational humour presents a uniquely demanding arena for model evaluation that complements the existing array of benchmarks. Lacking from existing options are foundational elements of humour, such as phonetic understanding and pragmatic inference. Consider the joke in which a husband and wife are solving a crossword. The joke centres on the husband giving clues such as "\textit{Emphatic no, five letters"}, and "\textit{Pistol, 3 letters}", resulting in guesses of "\textit{never}" and "\textit{gun}". This continues until the string of words guessed by the wife reads "never gun ugh give ewe Up". Whilst meaningless in isolation, the spoken realisation of the sequences equates to "\textit{Never gonna give you up}", an instance of the popular \textit{Rickrolling} internet phenomenon.\footnote{\href{https://www.youtube.com/watch?v=dQw4w9WgXcQ}{https://en.wikipedia.org/wiki/Rickrolling}}
The creation and comprehension of this joke depends on the recognition of Rick Astley's song of the same title, the cultural phenomenon of \textit{Rickrolling}, and an understanding of the phonetic similarity present in homophones.\footnote{This is made especially hard as there is not a 1-to-1 mapping, with "gun" and "ugh" taking the place of "gonna".}

Moreover, humour frequently depends on understanding what was \textit{not} said or inferring the opposite of explicit statements \cite{inbook}, which directly conflicts with the semantic-similarity-based retrieval systems that would typically be employed when addressing jokes that reference post-training events \cite{barnett2024sevenfailurepointsengineering, article_sem_sim}.

It is for these reasons that we assert the importance of humour understanding as a subdomain of NLP in the LLM age. The field of computational humour remains a fruitful area for continued work \cite{ignat-etal-2024-solved,lima-inacio-goncalo-oliveira-2023-towards}; one that is critically overlooked and under-researched.

\subsection{Theories of Humour}

Humour, as a uniquely human experience, has been the focus of thinkers for centuries, becoming the subject of heavy debate. As such, several prominent theories have arisen that attempt to explain the perception of humour, a subset of which are presented below:

\begin{itemize}
    \item \textbf{Relief Theory} suggests that humour, and particularly laughter, is the result of releasing psychological energy \cite{freud_relief_1963,Spencer,Kant1790}.
    \item \textbf{Superiority Theory} posits that the experience of humour and comedy is born out of the perception that one individual is superior to another, thus making the inferior individual the subject of humour \cite{hobbes_leviathan_1660,PlatoPhilebus, AristotlePoetics}.
    \item \textbf{Incongruity Theory} states that humour is the perception of something that conflicts with established mental patterns and expectations, therefore being incongruous \cite{morreal_2024, Tu2014}.
    \item \textbf{Benign Violation Theory} postulates that humour arises from situations that are simultaneously harmless (i.e., benign) and are incongruous with expectations (i.e., violating a norm). \cite{McGraw_BVT_2012,McGraw_BVT_2010}
\end{itemize}

Whilst the Superiority and Relief theories offer accounts of the experience of humour, they do not present simple interpretations that can be easily formulated linguistically and computationally. It is for this reason that, of the approaches to humour processing that are directly grounded in theory, incongruity and the general sense of norm violation remain essential elements \cite{tian-etal-2022-unified,he-etal-2019-pun,valitutti-etal-2013-everything}. 
We present examples of textual linguistic humour in \autoref{fig:text_examples} and an example of a multimodal humorous meme in \autoref{fig:meme}, to exemplify the broad range of possible humour forms.

\section{Humour Generation} 
\label{sec:generation}

The ability to compose novel jokes requires an implicit understanding of the cognitive mechanisms that underlie humour. As outlined in the \textit{Incongruity} and \textit{Benign Violation} theories of humour, a necessary feature of humorous language is that it violates an expectation within the reader, either in the form of cultural or situational norms (e.g., topical and contextual humour), or linguistic norms (e.g., puns). This core property poses a challenge to computational approaches and to creative language generation more broadly: the language modelling objective aims to maximise the log likelihood of an output sequence, thereby working against the very incongruity that humour demands \cite{train_and_constrain}. The following subsections explore historical attempts to address or incorporate this contention for pun generation.

\subsection{Pun Generation}
\label{sec:pun-generation}

Pun generation has dominated research, spanning from early rule-based approaches \cite{lessard_tom-swifties} to contemporary neural methods. The historical prevalence of puns is largely a result of their computational tractability and theoretical grounding in Incongruity Theory, providing both an entry point and capacity to produce large quantities of training/evaluation data. Early works, such as JAPE \cite{Binsted-JAPE-thesis,Binsted-Ritchie-JAPE-Conf} and STANDUP \cite{ritchie-STANDUP-2007}, leverage WordNet \cite{wordnet-hardcopy} for semantics and Unisyn\footnote{\url{https://www.cstr.ed.ac.uk/projects/unisyn/}.} for phonetic similarities. However, such classical approaches rely heavily on fixed schema, limiting overall creativity. For a deeper coverage of early literature in pun generation, we refer the reader to the prior surveys of \citet{ritchie_current_2001} and \citet{amin-burghardt-2020-survey}.

The remainder of this section is split according to the distinction of \textit{heterographic} and \textit{homographic} puns \cite{redfern_puns}, including their combination. See \autoref{fig:text_examples} for an example of each type.

\subsubsection{Homographic Pun Generation}
\label{sec:homographic}

Homographic puns exploit polysemy, the phenomenon of multiple meanings existing for a single word, to create humour through semantic ambiguity. Early work in automatic pun generation focused on this category due to the accessibility of semantic resources like WordNet \cite{miller-1994-wordnet}.

\citet{yu-etal-2018-neural} presents the first neural approach to homographic pun generation, training a conditional encoder-decoder LSTM on unlabelled Wikipedia text to create sentences that could support the semantics of two words simultaneously. This model then uses a constrained beam search algorithm to jointly decode the two distinct senses of the same word, generating puns without requiring any pun-specific training data.

In a follow-up work, \citet{yu-etal-2020-homophonic} explore pun generation through a lexically constrained rewriting approach that first identifies constraint words supporting semantic incongruity for a sentence, then rewrites it with explicit positive and negative constraints. Their method achieved state-of-the-art results in both automatic and human evaluations.

Building on this, \citet{luo-etal-2019-pun-GAN} introduced Pun-GAN, a GAN-based model \cite{goodfellow2014generativeadversarialnetworks} that employed a discriminator module with word-sense disambiguation capabilities to assess how well a generated sentence supported the polysemy of the target homographic pun word, aiming to maximise semantic ambiguity. The generator was trained via reinforcement learning using the discriminator's output as a reward signal to encourage the production of sentences that could support two word senses simultaneously. However, the key shortcoming of this approach was its tendency to produce generic outputs that prioritised semantic ambiguity over overall pun quality.

\textsc{AmbiPun} \cite{mittal-etal-2022-ambipun} takes as input a homograph with two distinct word senses and proposes that ambiguity comes from context rather than the pun word itself. The approach first produces a list of related concepts through a reverse dictionary, then utilises one-shot GPT-3 to generate context words from both concepts before generating puns that incorporate these contextual elements. They achieve a 52\% success rate in human evaluation, significantly outperforming baselines but still remaining tied to a complex, multi-staged pipeline.

\subsubsection{Heterographic Pun Generation}
\label{sec:heterographic}

Heterographic puns present additional challenges due to the requirement of modelling phonetic similarity between different surface forms while maintaining semantic coherence. These puns exploit words that sound alike but are spelt differently, requiring systems to understand both phonetic relationships and semantic contexts \cite{kao_puns}.

\citet{he-etal-2019-pun} generate heterographic puns through an unsupervised retrieve-and-edit framework based on the \textit{local-global surprisal} principle. Given a pair of homophones (e.g., "died" and "dyed"), they first retrieve candidate sentences containing the alternative word from a corpus, replace the alternative word with the pun word, and insert a semantically related topic word to the start of the pun. The approach is a direct instantiation of Incongruity Theory: the foreshadowing topic word creates a strong association with the pun word in the global context while maintaining the local context of the alternative word in the sentence.

\subsubsection{Combined Pun Generation}
\label{sec:combined}

More recently, \citet{tian-etal-2022-unified} proposed a unified framework that addresses both homographic and heterographic pun generation by incorporating three key linguistic attributes: ambiguity, distinctiveness, and surprise. Their approach demonstrates that principled integration of humour theories can improve generation quality across different pun types.

\citet{chen-etal-2024-u} propose a multi-stage curriculum learning approach using Direct Preference Optimisation (DPO) \cite{rafailov2024directpreferenceoptimizationlanguage} to align models with the ability to create valid linguistic structures in support of both homographic and homophonic pun creation. The authors ultimately released the \textit{ChinesePun} dataset of Chinese humour, presenting one of the few works that does not focus predominantly on English.

Recent systematic evaluation by \citet{xu-etal-2024-good} provides evidence that large language models struggle considerably more with the generation of heterographic puns than homographic puns. This difficulty likely stems from the need to infer phonetic characteristics of words, which is challenging for models operating primarily on text-based representations \cite{baluja-2025-text}. The authors additionally identify what they term "lazy pun generation," whereby models incorporate both senses of the pun word in a single generation (e.g., "The sailor's pay was docked after he struggled to dock on time"). This phenomenon nullifies the intended effects of surprisal, eliminating any ambiguity that would require cognitive effort to resolve.

\citet{CUP-puns} demonstrates a departure from standalone pun generation, releasing CUP: a novel task for \textit{context situated} puns. CUP integrates puns more naturally into real-world conversations by demanding contextual awareness. The authors extract keywords from context sentences using RAKE \cite{RAKE} and use a T5 model to create sentences that support both meanings of the pun word. Logically, this approach can be viewed as a hybrid of \citet{mittal-etal-2022-ambipun}, \citet{he-etal-2019-pun}, and \citet{luo-etal-2019-pun-GAN}.

\subsection{Non-Pun Humour Generation}
\label{sec:non-pun-generation}

Whilst puns have received the most attention in computational humour research, broader forms of humour generation present a greater challenge due to association with cultural knowledge, timeliness, and more complex incongruities. 

\citet{goel-etal-2024-automating} proposed an approach combining template extraction and infilling using BERT \cite{devlin2019bertpretrainingdeepbidirectional} with LLMs (specifically GPT-4, \citealp{openai2024gpt4technicalreport} and Zephyr-7B, \citealp{tunstall2023zephyrdirectdistillationlm}) to generate set-up and punchline style jokes (e.g., "\textit{Why did the chicken cross the road? To get to the other side}"). To achieve this, tokens from jokes are masked, and BERT attention weights are used to determine what elements of a given joke can be masked to maintain the essential structure, whilst removing overly topic-specific terms. As a result, the system learns to create joke structures that can then be filled to create novel joke instances.   

On the other hand, \citet{chung-etal-2024-visual} present \textsc{UNPIE}, a benchmark to assess the understanding of Vision Language Models (VLMs) when reasoning about lexical incongruities. To achieve this, they take as input written puns and generate an image to serve as a visual representation of the pun, demonstrating that such visual clues may be beneficial in pun understanding tasks, helping to identify the location of a pun in a given text.

\citet{horvitz-etal-2020-context} focus on the generation of satirical headlines, a type of language used that is both humorous and dry, being used to criticise individuals and entities on complex topics. In doing so, they tackle the linking of relevant real-world knowledge from sources such as Wikipedia and CNN, to relevant satirical headlines from TheOnion,\footnote{\url{https://theonion.com/}} which are then used to finetune GPT-2 \cite{GPT-2}.

\citet{tian-etal-2021-hypogen-hyperbole} explore humour through the lens of hyperbole with \textsc{HypoGen}. To do so, they curate a dataset of hyperbolic phrases following the "\textit{so [X] that [Y]}" pattern (e.g., "\textit{My personality is so dry that a cactus flourishes inside}"), using variants of  CoMET models \cite{bosselut-etal-2019-comet} to learn the commonsense and counterfactual relationships present in hyperbolic language to assist generation.

Finally, in recent years, there has been a growing body of work in the humour-adjacent domain of tongue twister generation (texts where entertainment and humour arise from mispronunciations stemming from complex phonetics). Such approaches involve training keyword-to-twister and style-transfer models for tongue twister generation, either via training on graphemes \cite{train_and_constrain,loakman-etal-2023-twistlist, keh-etal-2023-pancetta} or phonemes \cite{keh-etal-2023-pancetta}. Research in this domain has also highlighted the benefit of incorporating explicit phonetic and phonemic information into the generation of language formats that rely on such characteristics.

\section{Humour Explanation}
\label{sec:explanation}

\begin{displayquote}
"\textit{...if it can say why a joke's funny, it really does understand}" - Geoffrey Hinton.\footnote{From the 16th June 2025 episode of The Diary of a CEO podcast. See \url{https://youtu.be/giT0ytynSqg?si=O0iN3DM2Fv58fp8o&t=4460}.}
\end{displayquote}

The above quote from Hinton ("the Godfather of AI") refers directly to the use of joke explanation as a milestone achievement in the marketing of Google's PaLM LLM \cite{JMLR:v24:22-1144}. Detection approaches, while offering objective and easily automatable evaluation, remain susceptible to statistical flukes. Explanation generation eliminates this vulnerability through a vanishingly low probability of accidentally producing coherent comedic analysis. Moreover, the act of explanation provides valuable insight into the inner workings of what still exists as nearly opaque black boxes. This value comes at the expense of significantly more challenging, expensive and labour-intensive evaluation methodologies.

In its present state, the field of humour explanation remains critically understudied. To provide necessary context, we draw upon select literature covering humour classification and detection.

\subsection{Explanation through Classification}
\label{sec:classification}

Humour explanation, although understudied, is not a completely isolated evaluation of comprehension. Whilst not encompassing the full scope of a natural language explanation, humour \textit{detection} tasks models with identifying the linguistic traits common to humour. As such, we denote humour \textit{detection} a precursor to explanation.

The majority of existing works on humour explanation do not focus on providing textual natural language explanations, instead focusing on other indications of a joke's source, such as word senses or identifying the type of humour being expressed. 

\citet{miller-etal-2017-semeval} presents an overview of Task 7 from SemEval 2017, which concerned detection and "interpretation" (i.e., classification) of puns. Specifically, the pun interpretation task consisted of assigning word sense keys from WordNet \cite{wordnet-hardcopy} to the punning word contained in a text. One approach to the task, taken by \citet{oele-evang-2017-buzzsaw}, splits a given text into 2 parts at all possible locations, with the split where both parts have low semantic similarity being where the two meanings of the pun word are best separated. A Word Sense Disambiguation (WSD) model is then used to retrieve the relevant senses. Similar techniques based on mapping and comparing the semantics of different partitions of the text with the pun word are shown in multiple submissions (e.g., \citealp{hurtado-etal-2017-elirf, indurthi-oota-2017-fermi}). Whilst assigning sense keys to a pun word offers a minor explanation to a user, the requirement for the pun word to be pre-identified nullifies the appropriateness of such systems in practice.

\citet{PalmaPreciado2024JOKER} present an overview of the JOKER shared task from CLEF 2024, where Task 2 aimed to classify humorous texts based on their genre and the linguistic techniques used, including irony, sarcasm, exaggeration, incongruity/absurdity, self-deprecation and wit/surprise. In total, they received 54 submissions to the shared task, with techniques ranging from training BERT classifiers (e.g., \citealp{Narayanan2024CLEFJOKER}) and ensembles of classic machine learning classifiers (e.g., \citealp{Bartulovic2024LOLtoMDR}), to zero-shot prompting of LLMs like GPT-4 (e.g., \citealp{Wu2024HumourClassification} ).\footnote{See \citet{PalmaPreciado2024JOKER} for a full overview of submissions.}

However, explanation through classification presents a series of issues. Firstly, and most obviously, whilst such approaches can demonstrate a model's understanding of humour (to an extent), assigning the correct label to a given joke is unlikely to present a human user with valuable information that would aid their interpretation. Secondly, classification requires the creation of a taxonomy under which to categorise different jokes. Owing to the complexity of humour, high-quality, robust taxonomies are challenging to create. For instance, Task 2 from JOKER \cite{PalmaPreciado2024JOKER} has considerable overlap between joke categories. A prerequisite for both irony and sarcasm is the violation of a norm, whilst \textit{incongruity} is presented as a separate category. This is additionally true of "wit" (a subjective judgement of cleverness and novelty)  and \textit{surprise}.

\subsection{Natural Language Explanation}
\label{sec:natural_language_explanations}

Whilst explanation through the lens of classification is a valid approach for simple joke formats such as puns, more complex humour formats, such as more esoteric, context-dependent jokes, benefit from natural language explanations (in addition to being more friendly to the proposed end-user). Natural language explanations rectify the coarseness of explanation through classification. Such approaches do not require a pre-defined taxonomy of humour types to identify, but instead test the overall abilities of models to provide tailored, specific explanations for every humorous text.

\citet{xu-etal-2024-good} generates natural language explanations for puns using chain-of-thought prompting with a range of LLMs. They find that most LLMs are able to identify the punning word in both heterographic and homographic examples, but most models struggle with correctly identifying the alternative intended meaning of heterographic puns, which rely on phonetic similarity. The authors identify a series of mistakes commonly made by models, including failure to recognise the joke as a pun, incorrectly identifying the alternative senses of the pun word, and failing to provide the meanings of the pun word, but without contextualisation into a natural language explanation. 

\citet{loakman2025comparingapplesorangesdataset} extend these findings and use a range of LLMs on a range of joke formats, including homographic puns, heterographic puns, long-form humour, and topical jokes from Reddit. Their results further confirm that LLMs struggle with heterographic puns (owing to their lack of phonetic knowledge), but additionally demonstrate that longer incongruity-based jokes and jokes that rely on esoteric topical knowledge present even more difficulty to LLMs. Whilst a zero-shot non-chain-of-thought approach is used, the authors find that the Llama models distilled from Deepseek R1, are among the worst-performing. They hypothesise that this is a result of the ambiguity in real-world references within topical humour, leading to early misunderstandings being propagated through the reasoning process. A further, smaller-scale evaluation of ChatGPT's humour explanation ability is presented by \citet{jentzsch-kersting-2023-chatgpt}. Similarly, \citet{drivelology} investigate the ability of LLMs to explain "drivelology", a linguistic form characterised as "nonsense with depth", incorporating humour alongside other elements such as sarcasm, irony, and tautologies. Whilst they investigate the ability for LLMs to correctly categorise the type of "drivel" being used, they additionally perform zero-shot explanation generation and likewise find that models struggle significantly with creating high-quality explanations. 

Identifying the common failure of models to understand jokes where phonetic characteristics play a vital role (e.g., heterographic puns), \citet{baluja-2025-text} investigate whether access to speech audio of the joke being read aloud leads to improved performance. They assessed the ability of Gemini 1.5 \cite{geminiteam2024gemini15unlockingmultimodal} to explain jokes with and without access to text-to-speech readings, demonstrating approximately a 2.5\% to 4\% improvement in explanation performance. Whilst moderate, such findings indicate the affordances provided by multimodal prompting in language forms that rely heavily on modalities that are absent from text (i.e., pronunciations), suggesting room for further performance gains alongside improvements in the fusion of modalities.

\paragraph{Multimodal Humour and Meme Explanation}
In the realm of multimodal humour, \citet{hessel-etal-2023-androids} present work on the explanation of visual jokes in the form of humorous captions from the New Yorker Cartoon Caption Contest. In such a task, models must understand the visual cartoon in order to disambiguate pun words or correctly establish the key incongruity, owing to the text alone being ambiguous. For instance, one cartoon presents a barbershop with a hole in the roof and a spring coming out of a barbershop chair, with the caption reading "He'll be back". This phrase typically means that someone will return to an establishment even if they were dissatisfied (e.g., they provide an essential service or the individual's criticism was unreasonable). However, in this instance, the visual cues reveal that "He'll be back" is a literal statement, referring to the effects of gravity on the man who was ejected from his chair by a spring. To evaluate this, \citet{hessel-etal-2023-androids} finetune GPT-3 on human explanations, as well as perform 5-shot prompting with GPT-4. The results showed that access to visual information from the cartoons resulted in a better explanation in 84.7\% of cases (via human preference judgements). However, whilst in-context learning with 5-shot GPT-4 outperforms finetuned GPT-3, it was shown that human-authored explanations are still preferred in 68\% of instances. 

\begin{figure}
    \centering
    \includegraphics[width=0.8\linewidth]{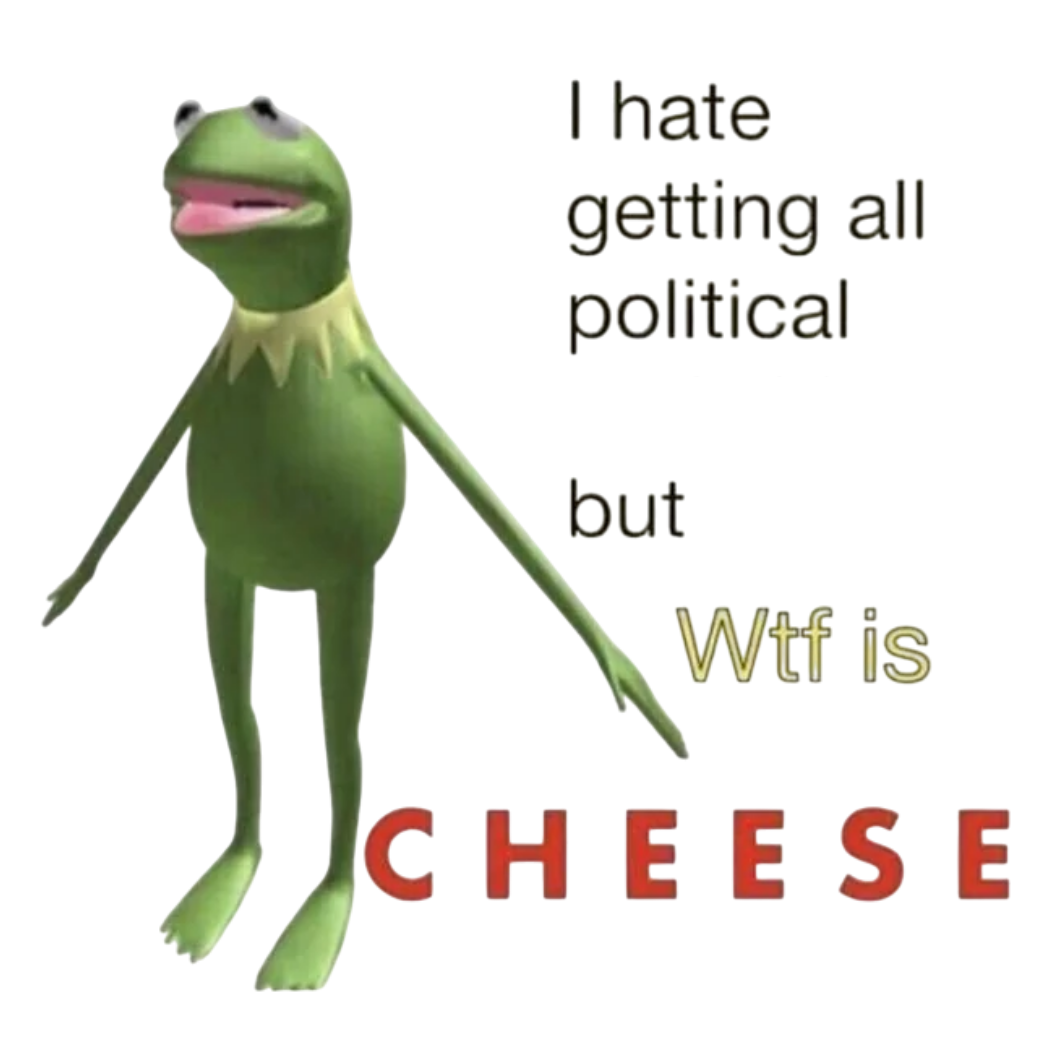}
    \caption{An example of a meme, a form of multimodal humour that incorporates visual elements, often alongside text. In this instance, the humour arises from the absurdity of the belief that asking about cheese is political, whilst being enhanced by the confused T-pose Kermit the Frog render. The original meme has been edited to remove expletives.}
    \label{fig:meme}
\end{figure}

On the other hand, the lion's share of work in multimodal humour explanation specifically investigates memes, a type of media (typically an image) that is copied and propagated rapidly across the internet, often comprising text captions and related imagery (see \autoref{fig:meme} for an example). \citet{hwang-shwartz-2023-memecap} present \textsc{MemeCap}, a dataset tailored to meme understanding, consisting of 6.3K memes from the \textit{r/Memes} subreddit alongside crowdsourced captions that explain the semantics that the meme is trying to convey. Additionally, \citet{khan-etal-2024-hope} present a dataset of 13K+ memes, including those with audio. In this instance, an LLM pipeline was utilised to generate explanations of the memes, which were then refined by human annotators. Interestingly, the authors aim to generate explanations that achieve the aim of entertaining and amusing readers (therefore being somewhat humorous themselves), rather than being strictly objective accounts of the humour.

\citet{park-etal-2024-memeintent} acknowledge that author's intent contributes significantly to meaning, ergo perception. They define the task of \textit{intent description generation}, accompanied by a dataset of 950 samples that are annotated with perceived intention and the necessary context. We believe that intent-aware systems are a compelling direction for future study, which we explore further in \S\ref{sec:intent-aware}.

Furthermore, \citet{agarwal-etal-2024-mememqa} present \textsc{MemeMQA}, a multimodal Q\&A dataset for asking questions regarding the content of memes in order to better understand them. From this, they develop \textsc{ARSENAL}, a multimodal meme understanding pipeline that uses LLMs to reason about a given meme in relation to a specific question.

The site \textit{Know Your Meme}\footnote{\url{https://knowyourmeme.com/}} tracks the provenance of memes as they develop, making it a wealth of knowledge regarding the origin and growth of memes over time. From this, \citet{tommasiniIMKGInternetMeme2023} built the Internet Meme Knowledge Graph, comprising 2 million edges to represent the semantics of multimodal memes.

In a more directly application-based setting, \citet{jha-etal-2024-meme} explore the explanation of memes being used explicitly for purposes of cyberbullying. Using the MultiBully dataset \cite{maity_2022}, they add annotations to highlight pertinent visual and linguistic aspects of a meme that highlight its intent as cyberbullying, allowing the training of enhanced bullying detection models.

\section{Discussion \& Future Directions}
\label{sec:discussion}

Generative tasks in the area of computational humour present a range of challenges owing to aspects such as the subjective and potentially offensive nature of humour and the ethics of generating any creative language form. In this section, we present a range of promising research directions for computational humour generation and explanation, relating each proposed direction to the practical and ethical challenges that they help ameliorate.

\subsection{Demographic Aware Humour Generation}
\label{sec:demographic_aware}

In following the recent trends established by works such as \citet{CUP-puns} and \citet{garimella-etal-2020-judge}, an essential primary focus of future research should be audience-tailored, demographic-aware, and contextually nuanced humour generation that is able to better address the preferences of particular end users. Owing to the wide gamut of topics that humour may be found in, such approaches would aid in decreasing the risk of generating material that is considered offensive to a given end-user. Consequently, a promising direction is to explicitly incorporate human preference alignment techniques such as RLHF \cite{ouyang_2022_rlhf, stiennon_2020_rlhf, christiano_2017)rlhf}, DPO \cite{rafailov_DPO_2023}, and PPO \cite{schulman2017proximalpolicyoptimizationalgorithms} to the task of humour generation.\footnote{See \citet{jiang2024surveyhumanpreferencelearning} for an overview of approaches to learning from human preference data.} 

Additionally, the developing area of perspectivist approaches in NLP \cite{fleisig-etal-2024-perspectivist,valette-2024-perspectivism,nlperspectives-2022-perspectivist} provide a route through which to consider individual preferences for subjective domains such as humour. Progress to this end is demonstrated by \citet{casola-etal-2024-multipico} and \citet{frenda-etal-2023-epic}, who present perspective-aware datasets for irony processing (a phenomenon highly related to humour). The types of humour and jokes that are most often required to be explained to someone are those that are ambiguous, nuanced, and frequently focused on sensitive topics. Whilst the potential sensitivity of a joke topic can have an intensifying impact on the level of humour perceived by the \textit{intended} audience, it acts as part of a risk-to-reward tradeoff, likewise increasing the risk of offence if presented to the wrong audience \cite{McGraw_BVT_2010}.

\subsection{Human-in-the-Loop Humour Generation} 
\label{sec:human-in-the-loop}

Humour is a quintessential demonstration of human intelligence and creativity. As such, the development of computational models for the generation of such content poses the risk of increasing the anthropomorphism of machine learning models - something that is at times desirable, but also disproportionately impacts vulnerable users of these technologies \cite{doi:10.55092/let20240003}. In addition to this, the creation of (intentional) humour is a creative endeavour. The generation of "\textit{creative}" language forms with computational models has the potential to have severe negative impacts on the arts and creative industries as a whole, and reduce the outlets available to people to realise financial gains from their own creativity. Whilst systems such as Witscript \cite{toplyn2023witscript3hybridai} were developed by real-world comedy writers, such examples are the exception to the rule. As a result, future work should focus increasingly on human-in-the-loop approaches \cite{WU2022364} to humour generation, requiring meaningful contributions from the end-user, or combining elements of humour generation and explanation to develop systems for joke workshopping, offering valuable feedback and direction on human-authored humour, rather than generating jokes wholesale.

\subsection{Complex Humour Explanation} 
\label{sec:complex_humour}

As highlighted in this survey paper, humour explanation is an interesting, challenging, and important task for assessing the verbal and commonsense reasoning abilities of models (particularly LLMs), yet is a scarcely researched domain. If the end goal is to imbue systems with human-level reasoning abilities, being able to explain humour is fundamental. Particular focus should be given to context-sensitive, esoteric humour such as topical jokes. Such jokes present a rich area for aiding in the development of advanced information retrieval systems that are able to work with ambiguous topics such as the references to world events and pop culture phenomena found in complex jokes.

\subsection{Intent-Aware Humour Explanation} 
\label{sec:intent-aware}
Whilst humour explanation is a challenging task with a real-world benefit of reducing communication barriers between people from different backgrounds, it is also an area with considerable ethical considerations. Another potential future direction for research in the area of humour explanation is the development of intent-aware models \cite{ma-etal-2025-detecting, park-etal-2024-memeintent}. Such approaches would attempt to model the characteristics and intent of the author based on available information (such as prior language use and known/inferred demographic variables) prior to attempting to explain potentially humorous language use. Not only could this information aid in inferring references to ambiguous content (aiding reasoning and relevant document retrieval), it would also aid in overcoming the undesirable effect of legitimising the instances where "\textit{it's just a joke!}" is used as a guise for spreading hatred through offensive material \cite{brommage2015just,hodson_joke_2010, FordRichardsonPetit+2015+171+186}. Such approaches have extensive applications in online communication venues such as social media and in handling interpersonal conflict. Whilst individual preferences and beliefs as to what topics are possible to derive humour from remain highly subjective \cite{smuts2010ethics, gaut1998just}, improved modelling of the underlying intent would help identify instances where offence was likely caused accidentally.\footnote{We of course do not advocate for the extreme version of such systems in practice, whereby individuals are effectively being accused of thought-crime due to an intent assigned to an author via a computational model.}

\section{Conclusion}
\label{sec:conclusion}

In this paper, we have explored the current landscape in the domain of computational humour generation and explanation, outlining the approaches taken in existing work and outlining promising future directions. We propose that computational humour processing remains an unsolved task with a wide range of real-world applications, as well as being one of the most promising yet overlooked domains in which to evaluate the verbal reasoning abilities of modern LLMs. Furthermore, we identified and outlined a range of future approaches to be taken in generative computational humour research and made the case that future work should focus both on an increased breadth of humour formats, as well as giving explicit consideration to the ethics of computational humour.

\section*{Limitations}

We position this work as an overview of computational humour generation and explanation. As a result, we focus on the breadth of research available, rather than presenting in-depth accounts of the exact technical novelty of existing works. Additionally, whilst we aim to be comprehensive with our overview, some instances of relevant research may have been missed. Furthermore, we have not extensively explored other areas of computational humour processing, such as automatic metrics and human evaluation paradigms, available datasets, or approaches to humour detection.

\section*{Ethics Statement}
Relating to the ethical considerations explored in this paper, whilst we believe that humour generation and explanation are worthwhile pursuits, we acknowledge and appreciate the stances taken by other individuals. This relates specifically to ethical considerations surrounding the generation of creative language in any form, the generation of humour that is potentially offensive, and whether or not providing an explanation for potentially offensive humour equates to an endorsement of the content of such humour. Such decisions should continue to be a source of discussion in the NLP community as a whole, and we encourage individual researchers working in these domains to explicitly state their stance in published works.

\subsubsection*{Acknowledgments}
Tyler Loakman is supported by the Centre for Doctoral Training in Speech and Language Technologies (SLT) and their Applications funded by UK Research and Innovation [grant number EP/S023062/1].

\bibliography{custom}

\end{document}